\documentclass[conference]{IEEEtran}
\IEEEoverridecommandlockouts
\usepackage{cite}
\usepackage{amsmath,amssymb,amsfonts}
\usepackage{algorithmic}
\usepackage{graphicx}
\usepackage{textcomp}
\usepackage{xcolor}
\def\BibTeX{{\rm B\kern-.05em{\sc i\kern-.025em b}\kern-.08em
    T\kern-.1667em\lower.7ex\hbox{E}\kern-.125emX}}
    
\usepackage{booktabs}
\usepackage{hyperref}
\usepackage{cleveref}
\pdfoutput=1

\begin{document}

\title{MultiScript30k: Leveraging Multilingual Embeddings to Extend Cross Script Parallel Data
}

\author{
    \IEEEauthorblockN{Christopher Driggers-Ellis}
    \IEEEauthorblockA{\textit{Computer and Information Science and Engineering} \\
    \textit{University of Florida}\\
    Gainesville, FL \\
    driggersellis.cw@ufl.edu}
    \and
    \IEEEauthorblockN{Detravious Brinkley}
    \IEEEauthorblockA{\textit{Computer and Information Science and Engineering} \\
    \textit{University of Florida}\\
    Gainesville, FL \\
    dj.brinkley@ufl.edu}
    \and
    \IEEEauthorblockN{Ray Chen}
    \IEEEauthorblockA{\textit{Computer and Information Science and Engineering} \\
    \textit{University of Florida}\\
    Gainesville, FL \\
    chenz1@ufl.edu}
    \and
    \IEEEauthorblockN{Aashish Dhawan}
    \IEEEauthorblockA{\textit{Computer and Information Science and Engineering} \\
    \textit{University of Florida}\\
    Gainesville, FL \\
    aashish.dhawan@ufl.edu}
    \and
    \IEEEauthorblockN{Daisy Zhe Wang}
    \IEEEauthorblockA{\textit{Computer and Information Science and Engineering} \\
    \textit{University of Florida}\\
    Gainesville, FL \\
    daisyw@ufl.edu}
    \and
    \IEEEauthorblockN{Christan Grant}
    \IEEEauthorblockA{\textit{Computer and Information Science and Engineering} \\
    \textit{University of Florida}\\
    Gainesville, FL \\
    christan@ufl.edu}
    }
\maketitle

\begin{abstract}
Multi30k is frequently cited in the multimodal machine translation (MMT) literature, offering parallel text data for training and fine-tuning deep learning models. 
However, it is limited to four languages: Czech, English, French, and German. 
This restriction has led many researchers to focus their investigations only on these languages.
As a result, MMT research on diverse languages has been stalled because the official Multi30k dataset only represents European languages in Latin scripts.
Previous efforts to extend Multi30k exist, but the list of supported languages, represented language families, and scripts is still very short.
To address these issues, we propose MultiScript30k, a new Multi30k dataset extension for global languages in various scripts, created by translating the English version of Multi30k (Multi30k-En) using NLLB200-3.3B.
The dataset consists of over \(30000\) sentences and provides translations of all sentences in Multi30k-En into Ar, Es, Uk, Zh\_Hans and Zh\_Hant.
Similarity analysis shows that Multi30k extension consistently achieves greater than \(0.8\) cosine similarity and symmetric KL divergence less than \(0.000251\) for all languages supported except Zh\_Hant which is comparable to the previous Multi30k extensions ArEnMulti30k and Multi30k-Uk.  
COMETKiwi scores reveal mixed assessments of MultiScript30k as a translation of Multi30k-En in comparison to the related work. ArEnMulti30k scores nearly equal MultiScript30k-Ar, but Multi30k-Uk scores $6.4\%$ greater than MultiScript30k-Uk per split.
\end{abstract}


\section{Introduction}
Multimodal machine learning has advanced in recent years, enabling innovative applications in various domains. 
However, the popular Multi30k dataset supports only the European languages Czech (Cs), English (En), French (Fr), and German (De) limiting linguistic diversity in multimodal machine translation (MMT) research and leaving many cultures and language families underrepresented.

This presents a significant gap in support for the intended MMT applications as described in the first Multi30k publication~\cite{elliott2016multi30k} because parallel text data in non-European languages is not included in Multi30k.
To bridge the gap, we propose the MultiScript30k extension of Multi30k to include Arabic (Ar), Spanish (Es), Ukrainian (Uk), and Chinese (Zh) via machine translation (MT) using the NLLB200-3.3B model~\cite{team2022NLLB} on the English Multi30k dataset (Multi30k-En)~\cite{elliott2016multi30k}. 

\begin{figure}
    \centering
    \includegraphics[width=1\linewidth]{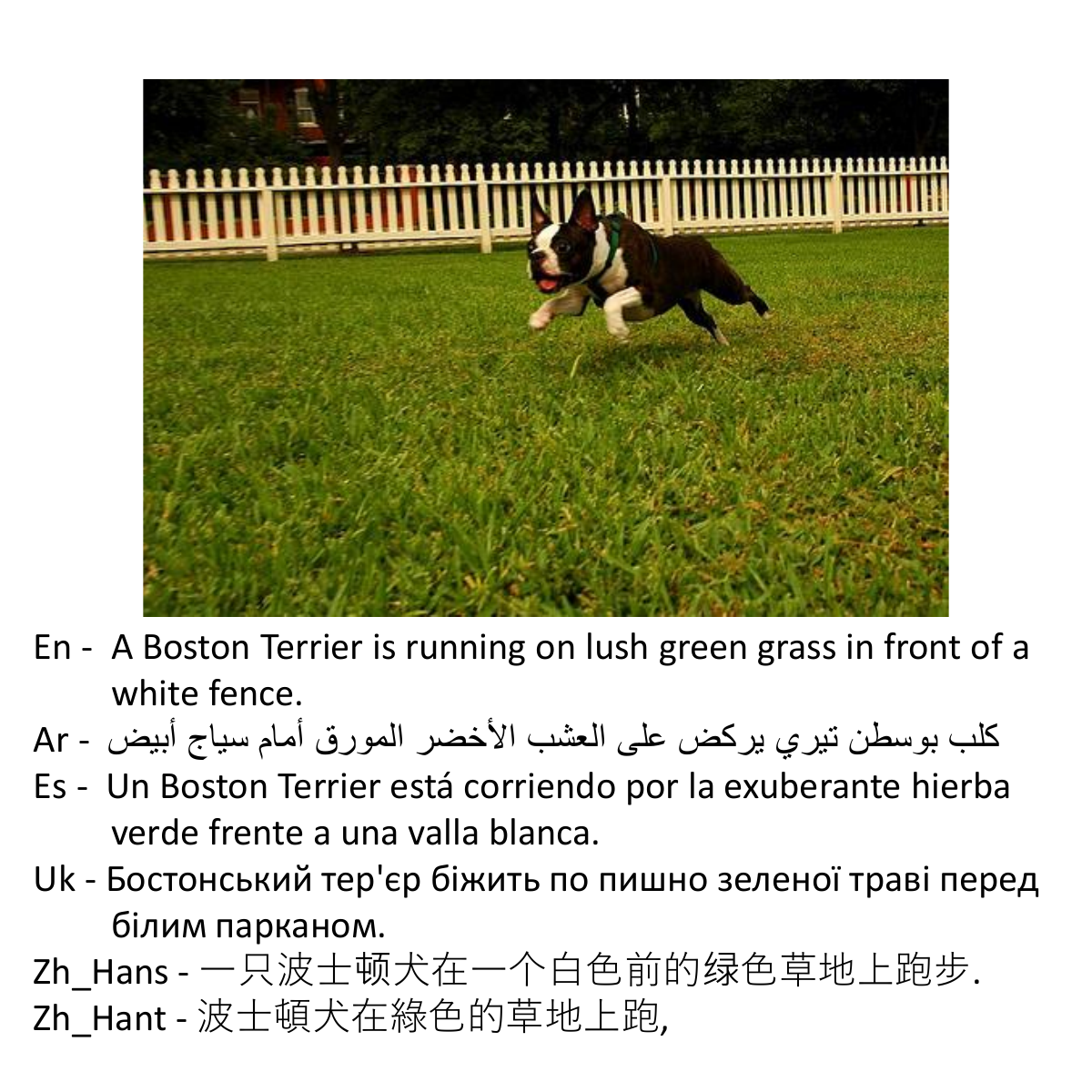}
    \caption{A record from the MultiScript30k dataset. The original English caption from Multi30k-En is the first entry. Other entries are the machine translations of the same caption which appear in MultiScript30k.}
    \label{fig:boston-terrier-translation}
\end{figure}

We evaluate the quality of translations using the widely adopted COMETKiwi metric, and through symmetric KL divergence and cosine similarity analysis on the multilingual text embeddings of original and translated text. 
In this manner, we find that the raw COMETKiwi scores for MultiScript30k as a translation of Multi30k-En are on par with winning submissions to the WMT24 general task on machine translation \cite{koc2024findings,rei2024tower} and that MultiScript30k is in semantic alignment with the original data and related work extending Multi30k.

Extending Multi30k is crucial for MMT research, enabling researchers to study the unique artifacts and challenges of MMT to and from the newly supported languages and scripts and to reach and impact the global communities that speak those languages.
By broadening the scope of multimodal datasets, our extension fosters more inclusive and diverse research in multimodal machine learning and facilitates deeper exploration of the MMT task.
Our contribution is the MultiScript30k.\footnote[1]{The MultiScript30k dataset is freely available and licensed under Creative Commons Attribution-NonCommercial-ShareAlike 4.0 International.
To view a copy of this license, visit \url{https://creativecommons.org/licenses/by-nc-sa/4.0/}. To access MultiScript30k, visit \url{https://github.com/ufdatastudio/multiscript30k}. Our generation and evaluation code are also available at \url{https://github.com/ufdatastudio/multi30k-extension}.} dataset, which extends the narrower Multi30k using MT as a synthetic data generation method. An example record from MultiScript30k is in~\Cref{fig:boston-terrier-translation}.

\section{Related Work}
MultiScript30k extends the popular MMT parallel text dataset Multi30k.
Multi30k was first introduced in \cite{elliott2016multi30k} as an En-De corpus made by translating En captions of images in Flickr30k~\cite{young2014image} to form En-De parallel data. 
The dataset was later extended in \cite{elliott2017findings} and \cite{barrault2018findings} to include Fr and Cs translations, respectively.
Multi30k-De, -En, and -Fr consist of four test splits (\emph{2016 Flickr}, \emph{2017 Flickr}, \emph{2017 MSCOCO}, and \emph{2018 Flickr}), one training split (\emph{train}), and one validation split (\emph{val}) while Multi30k-Cs does not have \emph{2017 Flickr} and \emph{2017 MSCOCO} test splits. 
Multi30k is very widely cited in the MMT literature.
It has been recognized as the most widely used dataset for training and fine-tuning models for the MMT task~\cite{caglayan2019probing, futeral2024zero, zhu2023beyond} and is also often used for benchmarking MMT approaches~\cite{gupta2023cliptrans, guo2022lvp, zhu2023beyond}. 

However, reliance on the Multi30k dataset is hindering the progress of MMT research because of the narrow range of languages it supports. 
All of these languages are European in origin and are written in Latin scripts. 
In particular, Chinese and Spanish, two of the most popular languages in the wider MT literature, are excluded.

The research community is also investigating methods to extend the Multi30k dataset.
~\cite{saichyshyna2023extension} uses Google Cloud Translate (GCT) to translate the Multi30k-En text into Ukrainian (Uk), human annotators to clean translations, and cosine similarity to evaluate translation quality by proxy. The result of their efforts was published as Multi30k-Uk.
The M\textsuperscript{3}-Multi30K dataset~\cite{guo2022lvp} uses cross-lingual representation learning model (XLM-R)~\cite{conneau2020unsupervised} for MT to extend Multi30k to Hindi, Latvian, and Turkish.
~\cite{mohammed2020aren} offers an Arabic translation of the Multi30k dataset called ArEnMulti30k, but the method used to generate this translation is unclear.



Several traditional metrics exist to compute the quality of MT. 
A more recent, popular and evolving metric is COMET and its derivatives ~\cite{rei2020comet, rei2022cometkiwi, rei2023scaling, rei2024tower}. 
This neural metric is built using cross-lingual pre-trained language modeling~\cite{rei2020comet} and is robust across various language pairs (such as En-Zh, En-Es, etc.) \cite{rei2022cometkiwi, rei2023scaling, rei2024tower} and domains (such as news, social networks, etc.)~\cite{rei2024tower}. 

Specifically, we use the COMETKiwi version of the COMET metric introduced in WMT23, which uses XLM-R XL as a pretrained encoder~\cite{rei2023scaling}. 
Our primary motivation is that it is a reference-free version of COMET. 
This enables us to measure MT quality of MultiScript30k without human-translated reference versions of the dataset in the target languages.
Preparing human translations/annotations would be prohibitively expensive and labor-intensive.
COMETKiwi has also become widely adopted among MT researchers.
It served as one of the primary automated metrics in the general MT shared task at WMT24~\cite{koc2024findings}, so using COMETKiwi as a metric not only enables reference-free translation research but provides a point of comparison to current MT systems, including high-performance systems like the winning WMT24 submission Tower-v2~\cite{rei2024tower}.


\begin{figure*}
    \centering
    \includegraphics[width=1\linewidth,height=0.5\linewidth,keepaspectratio]{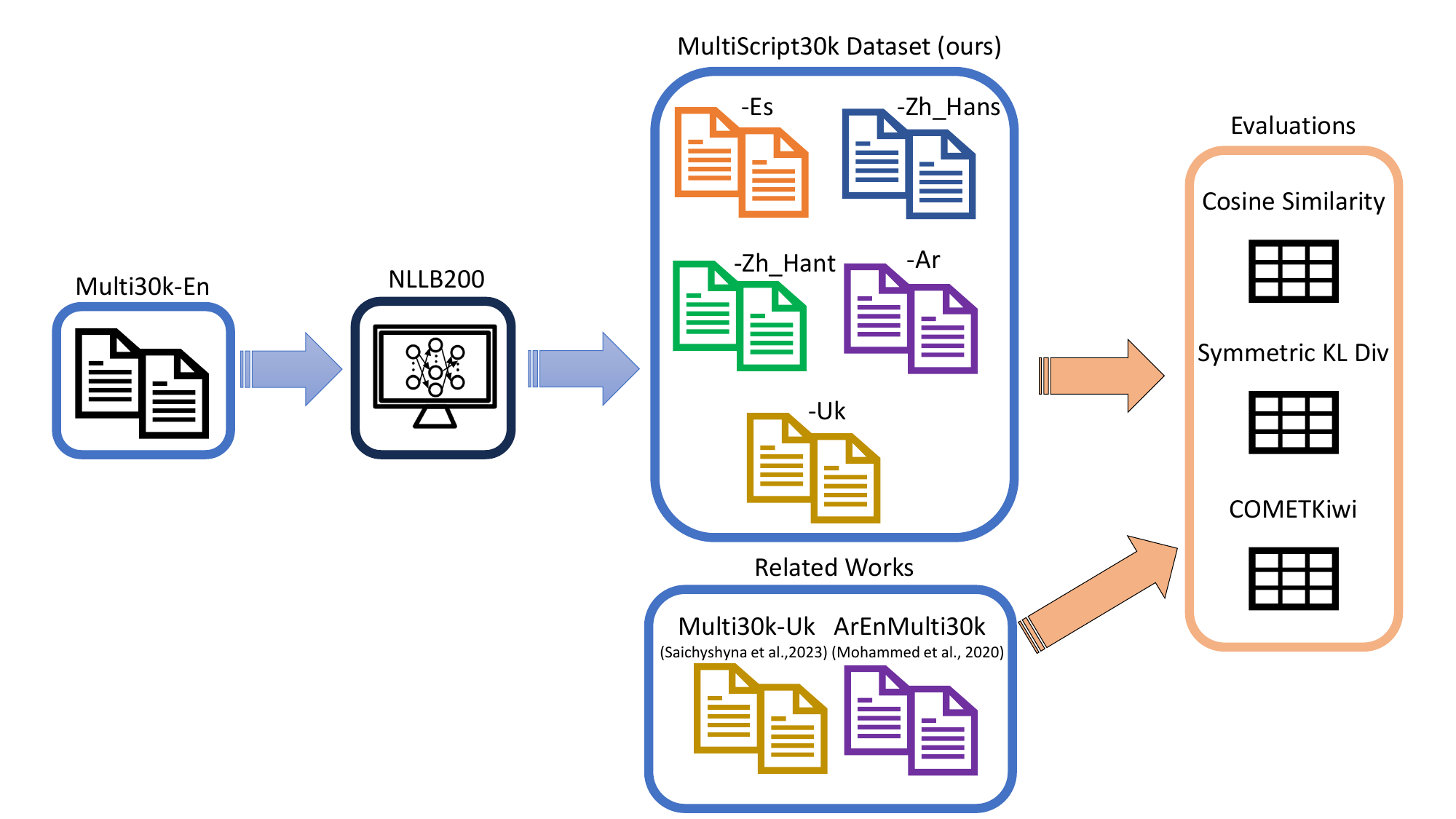}
    \caption{The workflow used to construct and evaluate the MultiScript30k dataset. \color{blue}Blue \color{black} Boxes represent data, the \textbf{Black} Box represents the NLLB200-3.3B MT model, and the \color{brown}Orange \color{black} Box represents evaluations. The original English data are translated using NLLB200-3.3 into 5 versions. Two datasets from related works on Multi30k extension are evaluated by way of comparison to MultiScript30k-Uk and MultiScript30k-Ar.}
    \label{fig:workflow}
\end{figure*}

\section{Methodology}
Our methodology for generating MultiScript30k is strongly inspired by~\cite{saichyshyna2023extension}. The authors outline a fast method by which a Multi30k dataset extension can be synthetically generated and assessed with or without human translators or evaluators. Additionally, the recently adopted COMETKiwi metric provides a ripe opportunity to evaluate Multi30k translations into various languages without the need of a reference translation.
We use MT to generate novel Multi30k extensions in Es and Zh and generate our own Uk and Ar datasets to enable comparison of our data to Multi30k-Uk~\cite{saichyshyna2023extension} and ArEnMulti30k~\cite{mohammed2020aren}
These additional languages promote diversity in the scripts and language morphology represented by MultiScript30k. 
The overall workflow is captured in~\Cref{fig:workflow}.

We select NLLB200-3.3B~\cite{team2022NLLB} as the MT method; the Multi30k-En data is translated into four target languages using the pretrained facebook/nllb-200-3.3B model available on HuggingFace.
We select the following languages for the initial MultiScript30k dataset: Arabic (Ar), Spanish (Es), Simplified Chinese (Zh\_Hans) and Traditional Chinese (Zh\_Hant), and Ukrainian (Uk).
The MT procedure creates a synthetically generated dataset in each of the target languages for all Multi30k-En splits whereas the previous Multi30k-Uk and ArEnMulti30k extensions lack translations of one or more of the splits. 
All translations were performed using 2 NVidia A100 GPUs, 16 CPUs and 64GBs of memory and default model parameters.

Since few native speakers are available to us for evaluation, we measure the cosine similarity and symmetric KL divergence of multilingual embedding vectors as a measure of semantic similarity between parallel sentences in Mutli30k-En and their translations. 
Following the example set by ~\cite{saichyshyna2023extension}, the HuggingFace sentence transformer distiluse-base-multilingual-cased-v2~\cite{reimers2019sentence-bert} is used to calculate multilingual embedding vectors for the original Multi30k-En, MultiScript30k translations, Multi30k-Uk and ArEnMulti30k.
Additionally, we use the large COMETKiwi model hosted at Unbabel/wmt23-cometkiwi-da-xl~\cite{rei2023scaling} to estimate the MT quality of both MutliScript30k and the previous Multi30k-Uk and ArEnMulti30k extension datasets as translations of Multi30k-En. 

For all data in MultiScript30k and the related work, cosine similarity and symmetric KL divergence between corresponding embedding vectors of the translations and Multi30k-En are calculated to compare the semantics of translations to the En source text. 
The results appear in \Cref{tab:embedding-space-comparison,tab:kl-div-comparison}, respectively. 
Cosine similarity and symmetric KL divergence are also calculated between the MultiScript30k-Uk data and the Multi30k-Uk dataset, as well as between MultiScript30k-Ar and ArEnMulti30k to semantically compare extensions in common languages prepared with differing methodologies.
These data appear in \Cref{tab:ar-metric-comparison,tab:uk-metric-comparison}.

Despite committing to cosine similarity and KL divergence analysis of embedding vectors as a measure of semantic similarity between MultiScript30k-Ar, MultiScript30k-Uk and the related work, little evidence was uncovered in the literature to explicitly link this metric to estimations of MT quality outside of the justification given in ~\cite{saichyshyna2023extension}.
Since previous extensions have already prepared translations of Multi30k-En into Ar~\cite{mohammed2020aren} and Uk~\cite{saichyshyna2023extension}, lexical BLEU and ChrF++ analysis is conducted on the applicable data using the SacreBLEU implementation~\cite{post2018call}. The related works act as reference translations for MultiScript30k-Ar and -Uk. Results are reported in \Cref{tab:ar-metric-comparison,tab:uk-metric-comparison}.

COMETKiwi~\cite{rei2023scaling} scores are calculated reference-free using unbabel-comet.
The original Multi30k-En is the source data and the translations to MultiScript30k, Multi30k-Uk and ArEnMulti30k are evaluated.
The results appear in~\Cref{tab:cometkiwi-comparison}.

\begin{table*}
\centering
\resizebox{\textwidth}{!}{%
\begin{tabular}{lccccccc}
\toprule
\textbf{Dataset} & \textbf{2016 Flickr} & \textbf{2017 Flickr} & \textbf{2017 MSCOCO} & \textbf{2018 Flickr} & \textbf{Train} & \textbf{Val} \\
\midrule
MultiScript30k-Ar & 0.8509 & 0.8389 & 0.8327 & 0.8483 & 0.8451 & 0.8477\\
MultiScript30k-Es & \textbf{0.9139} & \textbf{0.9103} & \textbf{0.9077} & \textbf{0.9167} & \textbf{0.9086} & \textbf{0.9114}\\
MultiScript30k-Uk & 0.8540 & 0.8433 & 0.8282 & 0.8545 & 0.8491 & 0.8496\\
MultiScript30k-Zh\_Hans & 0.8643 & 0.8513 & 0.8485 & 0.8460 & 0.8597 & 0.8598\\
MultiScript30k-Zh\_Hant & 0.7070 & 0.7222 & 0.7076 & 0.6858 & 0.6938 & 0.6922\\
Multi30k-Uk~\cite{saichyshyna2023extension} & 0.8487 & 0.8322 &	0.8413 & 0.8151 & 0.8428 & - \\
ArEnMulti30k~\cite{mohammed2020aren} & - & - & - & - & 0.8465 & 0.8473 \\
\bottomrule
\vspace{0.1pt}
\end{tabular}%
}
\caption{Mean cosine similarity of vector encoded MultiScript30k data, Multi30k-Uk~\cite{saichyshyna2023extension}, and ArEnMulti30k~\cite{mohammed2020aren} to Multi30k-En for all splits in the datasets. The bold indicates the dataset that has the highest similarity score for each Multi30k-En split in the embedding space.}
\label{tab:embedding-space-comparison}
\end{table*}

\begin{table*}
\centering
\label{tab:kl-div-comparison}
\resizebox{\textwidth}{!}{%
\begin{tabular}{lccccccc}
\toprule
\textbf{Dataset} & \textbf{2016 Flickr} & \textbf{2017 Flickr} & \textbf{2017 MSCOCO} & \textbf{2018 Flickr} & \textbf{Train} & \textbf{Val} \\
\midrule
MultiScript30k-Ar & 0.2330 & 0.2601 & 0.2656 & 0.2453 & 0.2408 & 0.2354 \\
MultiScript30k-Es & \textbf{0.1394} & \textbf{0.1501} & \textbf{0.1536} & \textbf{0.1397} & \textbf{0.1470} & \textbf{0.1421} \\
MultiScript30k-Uk & 0.2282 & 0.2530 & 0.2710 & 0.2361 & 0.2351 & 0.2330 \\
MultiScript30k-Zh\_Hans & 0.2154 & 0.2434 & 0.2445 & 0.2504 & 0.2220 & 0.2203 \\
MultiScript30k-Zh\_Hant & 0.4388 & 0.4278 & 0.4474 & 0.4805 & 0.4563 & 0.4578 \\
Multi30k-Uk~\cite{saichyshyna2023extension} & 0.2350 & 0.2685 & 0.2893 & 0.2554 & 0.2435 & - \\
ArEnMulti30k~\cite{mohammed2020aren} & - & - & - & - & 0.2391 & 0.2364 \\

\bottomrule
\vspace{0.1pt}
\end{tabular}%
}
\caption{Mean symmetric KL divergence (KL$\downarrow$) of vector encoded MultiScript30k data, Multi30k-Uk~\cite{saichyshyna2023extension}, and ArEnMulti30k~\cite{mohammed2020aren} from Multi30k-En for all splits in the datasets. The bold indicates the dataset that has the best score for each Multi30k-En split in the embedding space. \textbf{The values displayed represent KL divergence multiplied by $10^{3}$.}}
\label{tab:kl-div-comparison}
\end{table*}

\begin{table*}
\centering
\resizebox{\textwidth}{!}{%
\begin{tabular}{lccccccc}
\toprule
\textbf{Dataset} & \textbf{2016 Flickr} & \textbf{2017 Flickr} & \textbf{2017 MSCOCO} & \textbf{2018 Flickr} & \textbf{Train} & \textbf{Val} \\
\midrule
MultiScript30k-Ar & 0.7293 & 0.7334 & 0.7213 & 0.7124 & 0.7273 & 0.7297\\
MultiScript30k-Es & \textbf{0.7721} & \textbf{0.7811} & \textbf{0.7446} & \textbf{0.7593} & \textbf{0.7666} & \textbf{0.7591}\\
MultiScript30k-Uk & 0.7193 & 0.7042 & 0.6768 & 0.6877 & 0.7050 & 0.7105\\
MultiScript30k-Zh\_Hans & 0.6567 & 0.6226 & 0.6208 & 0.5947 & 0.6435 & 0.6438\\
MultiScript30k-Zh\_Hant & 0.4737 & 0.4819 & 0.4717 & 0.4161 & 0.4609 & 0.4577\\
Multi30k-Uk~\cite{saichyshyna2023extension} & 0.7582 & 0.7495 &	0.7362 & 0.7273 & 0.7450 & - \\
ArEnMulti30k~\cite{mohammed2020aren} & - & - & - & - & 0.7318 & 0.7313 \\
\bottomrule
\vspace{0.1pt}
\end{tabular}%
}
\caption{Mean COMETKiwi scores for MultiScript30k data, Multi30k-Uk~\cite{saichyshyna2023extension}, and ArEnMulti30k~\cite{mohammed2020aren} to Multi30k-En for all splits in the datasets. The bold indicates the dataset that has the highest COMETKiwi score on each Multi30k-En split.}
\label{tab:cometkiwi-comparison}
\end{table*}

\section{Results}
The results of our evaluations of MultiScript30k and the related works as translations of Multi30k-En are shown in~\Cref{tab:embedding-space-comparison,tab:kl-div-comparison,tab:cometkiwi-comparison}.
For each MultiScript30k translation, other than Traditional Chinese (Zh\_Hant), the mean cosine similarity for all splits of the Multi30k-En dataset are greater than \( 0.8\). 
Comparing MultiScript30k translations to each other, Spanish (Es) scores the highest mean cosine similarity for each of the data splits with more than \(0.9\) in each case.
Moving to comparisons between MultiScript30k and the related works' ArEn and Uk extensions, MultiScript30k-Ar achieves nearly equal similarity to Multi30k-En as the ArEnMulti30k~\cite{mohammed2020aren} dataset for the two splits available in that related work. 
We observe mixed differences in comparing MultiScript30k-Uk and Multi30k-Uk~\cite{saichyshyna2023extension}.
Differences are minimal, but MultiScript30k-Uk achieves slightly higher semantic similarity to the En source text for all represented splits in the original Multi30k dataset except \emph{2017 MSCOCO}.

~\Cref{tab:kl-div-comparison} displays the symmetric KL divergence of MultiScript30k and the related work from Multi30k-En.
Values are displayed at 1,000 times their true measurement for conciseness.

Among all datasets analyzed, MultiScript30k-Es consistently demonstrates the lowest (best) KL divergence values across all splits, scoring less than \(0.0002\) on every split, whereas all other datasets scored greater than \(0.0002\). This indicates that the Es dataset is the most aligned with the Multi30k-En in the multilingual embedding space, likely due to the structural and semantic similarities between the two languages.
In contrast, MultiScript30k-Zh\_Hant exhibits the highest (worst) KL divergences, consistently scoring above \(0.0004\) and roughly triple the KL divergence achieved by MultiScript30k-Es.
This higher divergence is likely due to the linguistic and syntactic differences between En and Zh\_Hant. 
Furthermore, the limited availability of high-quality training data between En and Zh\_Hant, as compared to Simplified Chinese (Zh\_Hans), may contribute to the disparity between En-Zh\_Hant and En-Zh\_Hans divergence.

For all datasets, the KL divergence values are stable across splits, as best observed in MultiScript30k-Es. 
This consistency indicates that the embedding space generalizes well across different subsets of the original Multi30k-En data.
The results for Multi30k-Uk and ArEnMulti30k suggest similar alignment with Multi30k-En as compared to their MultiScript30k counterparts. 
Thus, these datasets provide alternative multilingual resources with similar levels of compatibility with the En source text.

Results of COMETKiwi analysis on the MultiScript30k dataset and the related works are presented in ~\Cref{tab:cometkiwi-comparison}. 
In agreement with our cosine similarity and symmetric KL divergence analyses, MultiScript30k-Es achieves the greatest COMETKiwi score out of any Multi30k extension studied for every split in the original dataset. 
Likewise, MultiScript30k-Zh\_Hans and MultiScript30k-Zh\_Hant show large deficits in COMETKiwi scores as compared to MultiScript30k in other languages and the related work, and scores for MultiScript30k-Zh\_Hant are especially low.
Unlike cosine similarity and KL divergence analysis, the comparison between COMETKiwi scores for MultiScript30k and the related works differs greatly based on language.
Scores for MultiScript30k-Ar and ArEnMulti30k \emph{train} differ by only \(0.0045\), and scores for the \emph{val} splits are also very similar, differing by just \(0.0016\).
Multi30k-Uk consistently outperforms MultiScript30k-Uk by a wider margin.
The related work's COMETKiwi score for each split is \(0.045\) or $6.4\%$ greater on average. 

\begin{table} 
    \centering
    \resizebox{0.4\textwidth}{!}{
    \begin{tabular}{ccccc}
    \toprule
       \textbf{Split}  & \textbf{BLEU} & \textbf{ChrF++} & \textbf{CoSim} & \textbf{KL}$\downarrow *10^-3$ \\
    \midrule
        \textbf{Train} & 36.7 & 63.5 & 0.9234 & 0.09502 \\
        \textbf{Val}   & 38.2 & 64.8 & 0.9230 & 0.09573 \\
    \bottomrule
    \end{tabular}%
    }
    \caption{BLEU, ChrF++, cosine similarity (CoSim), and symmetric KL divergence (KL$\downarrow$) scores of MultiScript30-Ar, treating ArEnMulti30k~\cite{mohammed2020aren} as a reference translation. \textbf{The values displayed represent KL divergence multiplied by $10^{3}$.}}
    \label{tab:ar-metric-comparison}
\end{table}

\begin{table}
    \centering
    \resizebox{0.45\textwidth}{!}{
    \begin{tabular}{ccccc}
    \toprule
        \textbf{Split} & \textbf{BLEU} & \textbf{ChrF++} & \textbf{CoSim} & \textbf{KL}$\downarrow *10^-3$ \\ 
    \midrule
         \textbf{2016 Flickr} & 37.10 & 65.36 & 0.9243 & 0.09517\\
         \textbf{2017 Flickr} & 34.05 & 62.36 & 0.9078 & 0.1099\\
         \textbf{2017 MSCOCO} & 33.14 & 63.31 & 0.9050 & 0.1070\\
         \textbf{2018 Flickr} & 34.14 & 62.81 & 0.9187 & 0.1010\\
         \textbf{Train} & 35.77 & 63.59 & 0.9191 & 0.1006\\
    \bottomrule
    \end{tabular}
    }
    \caption{BLEU, ChrF++, cosine similarity (CoSim), and symmetric KL divergence (KL$\downarrow$) scores of MultiScript30k-Uk, treating Multi30k-Uk~\cite{saichyshyna2023extension} as a reference translation. \textbf{The values displayed represent KL divergence multiplied by $10^{3}$.}}
    \label{tab:uk-metric-comparison}
\end{table}
 
\Cref{tab:ar-metric-comparison,tab:uk-metric-comparison} convey the results of BLEU and ChrF++ and cosine/KL divergence similarity analysis of MultiScript30k-Ar and -Uk, treating ArEnMulti30k and Multi30k-Uk as reference translations.
Across all splits in both languages, the distribution of both metrics is narrow.
The BLEU metric's range is between \(33.14\) and \(38.2\), and ChrF++ ranges between \(62.36\) and \(65.36\).
Semantic similarity analysis of embedding vectors is high and also distributed narrowly between \(0.90\) and \(0.93\) (maximum is \(1\)) for Uk and between \(0.923\) and \(0.924\) for Ar. 
As would be expected, KL divergence for both \emph{train} and \emph{val} in the Ar data from ArEnMulti30k shown in ~\Cref{tab:ar-metric-comparison} is less than $50\%$ of the KL divergence from the Multi30k-En source text.
A similar result appears in ~\Cref{tab:uk-metric-comparison} for the Uk datasets.

\section{Discussion}
MultiScript30k-Es's high performance compared to other translations is in line with traditional expectations for MT tasks. 
Spanish (Es) is by far the most similar language to English (En) out of those studied, and the cosine similarity, symmetric KL divergence and COMETKiwi results for Es data are all better than the other languages studied, reflecting the challenge of MT to structurally diverse languages.
The Traditional Chinese (Zh\_Hant) dataset MultiScript30k-Zh\_Hant performed worst out of the translations studied in all three metrics.

Our investigation lends potential quantitative support to the cosine similarity metric as it agrees with other similarity measures. For each data split of the data compared, the cosine similarity metric remains within a narrow range, like the similar embedding-dependent symmetric KL divergence and the lexical substring similarity metrics BLEU and ChrF++. 
The BLEU and ChrF++ scores gathered in the secondary investigation stand as measures of MultiScript30k-Ar and -Uk's lexical similarity to the related work.

In the more direct COMETKiwi assessment of MT quality, however, MultiScript30k-Uk is outperformed by Multi30k-Uk by $6.4\%$ on average while MultiScript30k-Ar performs almost identically to ArEnMulti30k.
This finding suggests that the related work by ~\cite{saichyshyna2023extension} is superior to MultiScript30k for the Uk language it supports.
This finding quantitatively affirms the expectation that the workflow elaborated by ~\cite{saichyshyna2023extension}, involving human evaluators, will produce superior translations to unrefined synthetic MT data without post-processing.
Simultaneously, the average shortcoming of $6.4\%$ in COMETKiwi performance of non-annotated synthetic data could be tolerable for certain use cases if translations must be prepared rapidly or at little to no cost.

Although the data was generated using a different MT method (NLLB v GCT), our Ukrainian translation was very similar to those of Multi30k-Uk, yet the palpable difference in COMETKiwi evaluation suggests that whatever discrepancies exist in MultiScript30k-Uk are a detriment to its performance as an MMT dataset.
The same cannot be said for MultiScript30k-Ar in comparison to ArEnMulti30k.
The related work by ~\cite{mohammed2020aren} scored similarly in all three metrics to MultiScript30k-Ar so that whatever discrepancies exist in the two datasets resulted in minimal differences in their quality as translations of Multi30k-En.

Being optimistic, we draw a line of comparison between the COMETKiwi analysis and the data provided in the results of the WMT24 shared task on MT~\cite{koc2024findings}.
For every language pair studied, the findings of the WMT24 shared task declare Unbabel's Tower-v2 the most performant submission in its automatic metric rankings~\cite{koc2024findings,rei2024tower}. 
Tower-v2 was also the top submission in a human evaluation ranking for eight language pairs out of eleven~\cite{koc2024findings,rei2024tower}.
~\cite{rei2024tower} reports \(0.745\) and \(0.732\) COMETKiwi scores for the En-Es and En-Uk language pairs on the WMT24 evaluation data, respectively.
Though the data used are undoubtedly different, for each split of Multi30k except \emph{2017 MSCOCO}, COMETKiwi evaluation of MultiScript30k-Es scores its quality slightly higher than En-Es translations produced by the state-of-the-art Tower-v2 system.

In this work, we have presented the MultiScript30k dataset and demonstrated the viability of using MT for generating synthetic data to extend Multi30k-En by producing translations in several target languages that were both semantically similar to the source text and previous extensions.
These translations also scored well in COMETKiwi evaluations.
By leveraging the NLLB200-3.3B model for MT, we were able to create new parallel text data for Ar, Es, Uk and Zh.
The results indicate that these synthetic datasets are good translations and maintain semantic alignment with their En source text, making them potentially useful resources for MMT research. However, MultiScript30k-Ar and MultiScript30k-Uk are matched or outperformed by their counterparts in related works.

Despite the high cosine similarity scores, a native Zh speaker and collaborator in this research evaluated small samples of the Zh\_Hans and Zh\_Hant translations and found that the grammar in these translations was poor. 
This is in keeping with poor COMETKiwi scores produced for these translations and reveals a potential shortcoming of semantic similarity analysis for MT tasks.

Overall, findings support the use of synthetic data generation via MT but only in scenarios where human translators/annotators are unavailable or rapid dataset creation is absolutely necessary.
The MT quality of MultiScript30k is consistently below a previous synthetic Multi30k extension where human evaluators were used.
Synthetic data generation for MMT allows for the expansion of datasets such as Multi30k to include a more diverse array of languages whenever the temporal and/or fiscal overheads of human translation and/or annotation would be prohibitive, thereby promoting more inclusive and comprehensive MMT research in a variety of language families and scripts.

\section{Limitations}
Despite the general success in generating accurate translations, there are limitations and caveats surrounding MultiScript30k.
Although synthetic data generation shows promise as a method for addressing data scarcity, we cannot recommend using MultiScript30k as a serious replacement for human translation data in MMT research or in application at this time.

While the translations generally captured the correct meaning, a spot check of MultiScript30k-Zh\_Hans and -Zh\_Hant revealed instances where the grammar of the translations was suboptimal.
This issue highlights the challenges of relying solely on MT for high-quality text generation, particularly for languages with complex, differing grammatical structures or less representation in training data for the MT method.
It appears that syntax and grammar are ignored in favor of the meaning in the multilingual embedding cosine similarity metric whereas this problem evidently does not exist or is much less severe in COMETKiwi.
While MultiScript30k expands the Multi30k dataset to include various diverse languages, only a minimal manual review has been performed for one target language.
Most of MultiScript30k has yet to receive any native speaker evaluation, raising concerns about the understandability of the translations despite high COMETKiwi results for most languages.

MultiScript30k-Es has not been compared to another extension of Multi30k into the Es language because no such translation could be found in a literature search prior to our experiments.
No human evaluation has been performed on MultiScript30k-Es either.
Because there is no clear point of comparison, it is uncertain whether MultiScript30k-Es provides a good substitute for human translated data despite its performance in the quantitative analysis.
MultiScript30k-Es outscored Tower-v2 in En-Es translation quality as measured by COMETKiwi, but these scores were recorded for translations of different source data.
Only conventional wisdom suggests that MultiScript30k-Es is saved from the grammatical issues persistent in its Zh counterparts, due to the structural similarities between En and Es, whereas no evidence has been produced at present to support that hypothesis.

Finally, for Ar and Uk, MultiScript30k was outperformed or only matched the related work's COMETKiwi scores.
This fact suggests our workflow for generating the dataset needs improvement.

\section{Future Work}

Future work should focus on expanding the range of languages and language families in Multi30k extensions as well as improving the translation quality of synthetically generated datasets. 

Foremost, the MT method and overall workflow used to generate datasets should be improved compared to the present MultiScript30k dataset. Human evaluators who speak the target languages supported by the synthetic dataset should review and correct the data whenever available. 
This was the approach taken by ~\cite{saichyshyna2023extension} in making Multi30k-Uk, and that dataset outperformed MultiScript30k-Uk substantially.

Further research is needed to assess the validity of semantic similarity metrics, such as cosine similarity and symmetric KL divergence, as proxy measures of MT quality. 
Given that cosine similarity was only slightly decreased for MultiScript30k-Zh\_Hant and that the difference was much more pronounced for KL divergence and COMETKiwi, there is motivation for future works comparing semantic similarity metrics to human evaluations of MT quality to gain a more nuanced understanding of how they relate to MT performance.

Generally, future investigations should aim to refine synthetic data generation methodologies and metrics, enhancing the quality and selection of multimodal datasets for diverse linguistic and cultural contexts.
Only more and better multimodal data will improve data availability for multimodal machine learning tasks in diverse language families and scripts.

\section{Acknowledgments}
The summary of the COMETKiwi metric in Section 2 was written with inputs from Microsoft CoPilot. Text generated from the model was checked against the cited publications for accuracy and paraphrased.

\bibliography{citations}

@inproceedings{barrault2018findings,
  title={Findings of the Third Shared Task on Multimodal Machine Translation},
  author={Barrault, Lo{\"\i}c and Bougares, Fethi and Specia, Lucia and Lala, Chiraag and Elliott, Desmond and Frank, Stella},
  booktitle={Proceedings of the Third Conference on Machine Translation: Shared Task Papers},
  pages={304--323},
  year={2018}
}

@inproceedings{caglayan2019probing,
    title = "Probing the Need for Visual Context in Multimodal Machine Translation",
    author = {Caglayan, Ozan  and
      Madhyastha, Pranava  and
      Specia, Lucia  and
      Barrault, Lo{\"\i}c},
    editor = "Burstein, Jill  and
      Doran, Christy  and
      Solorio, Thamar",
    booktitle = "Proceedings of the 2019 Conference of the North {A}merican Chapter of the Association for Computational Linguistics: Human Language Technologies, Volume 1 (Long and Short Papers)",
    month = jun,
    year = "2019",
    address = "Minneapolis, Minnesota",
    publisher = "Association for Computational Linguistics",
    url = "https://aclanthology.org/N19-1422",
    doi = "10.18653/v1/N19-1422",
    pages = "4159--4170",
}

@inproceedings{conneau2020unsupervised,
    title = "Unsupervised Cross-lingual Representation Learning at Scale",
    author = "Conneau, Alexis  and
      Khandelwal, Kartikay  and
      Goyal, Naman  and
      Chaudhary, Vishrav  and
      Wenzek, Guillaume  and
      Guzm{\'a}n, Francisco  and
      Grave, Edouard  and
      Ott, Myle  and
      Zettlemoyer, Luke  and
      Stoyanov, Veselin",
    editor = "Jurafsky, Dan  and
      Chai, Joyce  and
      Schluter, Natalie  and
      Tetreault, Joel",
    booktitle = "Proceedings of the 58th Annual Meeting of the Association for Computational Linguistics",
    month = jul,
    year = "2020",
    address = "Online",
    publisher = "Association for Computational Linguistics",
    url = "https://aclanthology.org/2020.acl-main.747",
    doi = "10.18653/v1/2020.acl-main.747",
    pages = "8440--8451",
}

@inproceedings{elliott2016multi30k,
  author = 	"Elliott, Desmond
		and Frank, Stella
		and Sima'an, Khalil
		and Specia, Lucia",
  title = 	"Multi30K: Multilingual English-German Image Descriptions",
  booktitle = 	"Proceedings of the 5th Workshop on Vision and Language",
  year = 	"2016",
  publisher = 	"Association for Computational Linguistics",
  pages = 	"70--74",
  location = 	"Berlin, Germany",
  doi = 	"10.18653/v1/W16-3210",
  url = 	"http://www.aclweb.org/anthology/W16-3210"
}

@inproceedings{elliott2017findings,
  author    = {Elliott, Desmond  and  Frank, Stella  and  Barrault, Lo\"{i}c  and  Bougares, Fethi  and  Specia, Lucia},
  title     = {Findings of the Second Shared Task on Multimodal Machine Translation and Multilingual Image Description},
  booktitle = {Proceedings of the Second Conference on Machine Translation, Volume 2: Shared Task Papers},
  month     = {September},
  year      = {2017},
  address   = {Copenhagen, Denmark},
  publisher = {Association for Computational Linguistics},
  pages     = {215--233},
  url       = {http://www.aclweb.org/anthology/W17-4718}
}

@misc{futeral2024zero,
      title={Towards Zero-Shot Multimodal Machine Translation}, 
      author={Matthieu Futeral and Cordelia Schmid and Beno{\^\i}t Sagot and Rachel Bawden},
      year={2024},
      eprint={2407.13579},
      archivePrefix={arXiv},
      primaryClass={cs.CL},
      url={https://arxiv.org/abs/2407.13579}, 
}

@inproceedings{guo2022lvp,
    title = "{LVP}-{M}3: Language-aware Visual Prompt for Multilingual Multimodal Machine Translation",
    author = "Guo, Hongcheng  and
      Liu, Jiaheng  and
      Huang, Haoyang  and
      Yang, Jian  and
      Li, Zhoujun  and
      Zhang, Dongdong  and
      Cui, Zheng",
    editor = "Goldberg, Yoav  and
      Kozareva, Zornitsa  and
      Zhang, Yue",
    booktitle = "Proceedings of the 2022 Conference on Empirical Methods in Natural Language Processing",
    month = dec,
    year = "2022",
    address = "Abu Dhabi, United Arab Emirates",
    publisher = "Association for Computational Linguistics",
    url = "https://aclanthology.org/2022.emnlp-main.184",
    doi = "10.18653/v1/2022.emnlp-main.184",
    pages = "2862--2872",
}

@inproceedings {gupta2023cliptrans,
    title={CLIPTrans: Transferring Visual Knowledge with Pre-trained Models for Multimodal Machine Translation},
    author={Gupta, Devaansh and Kharbanda, Siddhant and Zhou, Jiawei and Li, Wanhua and Pfister, Hanspeter and Wei, Donglai},
    booktitle={Proceedings of the IEEE/CVF International Conference on Computer Vision},
    year={2023}
}

@inproceedings{koc2024findings,
    title = "Findings of the {WMT}24 General Machine Translation Shared Task: The {LLM} Era Is Here but {MT} Is Not Solved Yet",
    author = "Kocmi, Tom  and
      Avramidis, Eleftherios  and
      Bawden, Rachel  and
      Bojar, Ond{\v{r}}ej  and
      Dvorkovich, Anton  and
      Federmann, Christian  and
      Fishel, Mark  and
      Freitag, Markus  and
      Gowda, Thamme  and
      Grundkiewicz, Roman  and
      Haddow, Barry  and
      Karpinska, Marzena  and
      Koehn, Philipp  and
      Marie, Benjamin  and
      Monz, Christof  and
      Murray, Kenton  and
      Nagata, Masaaki  and
      Popel, Martin  and
      Popovi{\'c}, Maja  and
      Shmatova, Mariya  and
      Steingr{\'\i}msson, Steinth{\'o}r  and
      Zouhar, Vil{\'e}m",
    editor = "Haddow, Barry  and
      Kocmi, Tom  and
      Koehn, Philipp  and
      Monz, Christof",
    booktitle = "Proceedings of the Ninth Conference on Machine Translation",
    month = nov,
    year = "2024",
    address = "Miami, Florida, USA",
    publisher = "Association for Computational Linguistics",
    url = "https://aclanthology.org/2024.wmt-1.1",
    doi = "10.18653/v1/2024.wmt-1.1",
    pages = "1--46",
}

@misc{mohammed2020aren,
    author="Mohammed, Roweida and Inad Aljarrah, Mahmoud Al-Ayyoub and Ali Fadel",
    title="Arenmulti30k",
    year={2020},
    month=dec,
    publisher="Zenodo",
    url="https://doi.org/10.5281/zenodo.4394718",
}

@inproceedings{post2018call,
  title = "A Call for Clarity in Reporting {BLEU} Scores",
  author = "Post, Matt",
  booktitle = "Proceedings of the Third Conference on Machine Translation: Research Papers",
  month = oct,
  year = "2018",
  address = "Belgium, Brussels",
  publisher = "Association for Computational Linguistics",
  url = "https://www.aclweb.org/anthology/W18-6319",
  pages = "186--191",
}

@inproceedings{rei2020comet,
    title = "{COMET}: A Neural Framework for {MT} Evaluation",
    author = "Rei, Ricardo  and
      Stewart, Craig  and
      Farinha, Ana C  and
      Lavie, Alon",
    editor = "Webber, Bonnie  and
      Cohn, Trevor  and
      He, Yulan  and
      Liu, Yang",
    booktitle = "Proceedings of the 2020 Conference on Empirical Methods in Natural Language Processing (EMNLP)",
    month = nov,
    year = "2020",
    address = "Online",
    publisher = "Association for Computational Linguistics",
    url = "https://aclanthology.org/2020.emnlp-main.213",
    doi = "10.18653/v1/2020.emnlp-main.213",
    pages = "2685--2702",
}

@inproceedings{rei2022cometkiwi,
    title = "{COMET}-22: Unbabel-{IST} 2022 Submission for the Metrics Shared Task",
    author = "Rei, Ricardo  and
      C. de Souza, Jos{\'e} G.  and
      Alves, Duarte  and
      Zerva, Chrysoula  and
      Farinha, Ana C  and
      Glushkova, Taisiya  and
      Lavie, Alon  and
      Coheur, Luisa  and
      Martins, Andr{\'e} F. T.",
    editor = {Koehn, Philipp  and
      Barrault, Lo{\"\i}c  and
      Bojar, Ond{\v{r}}ej  and
      Bougares, Fethi  and
      Chatterjee, Rajen  and
      Costa-juss{\`a}, Marta R.  and
      Federmann, Christian  and
      Fishel, Mark  and
      Fraser, Alexander  and
      Freitag, Markus  and
      Graham, Yvette  and
      Grundkiewicz, Roman  and
      Guzman, Paco  and
      Haddow, Barry  and
      Huck, Matthias  and
      Jimeno Yepes, Antonio  and
      Kocmi, Tom  and
      Martins, Andr{\'e}  and
      Morishita, Makoto  and
      Monz, Christof  and
      Nagata, Masaaki  and
      Nakazawa, Toshiaki  and
      Negri, Matteo  and
      N{\'e}v{\'e}ol, Aur{\'e}lie  and
      Neves, Mariana  and
      Popel, Martin  and
      Turchi, Marco  and
      Zampieri, Marcos},
    booktitle = "Proceedings of the Seventh Conference on Machine Translation (WMT)",
    month = dec,
    year = "2022",
    address = "Abu Dhabi, United Arab Emirates (Hybrid)",
    publisher = "Association for Computational Linguistics",
    url = "https://aclanthology.org/2022.wmt-1.52",
    pages = "578--585",
}

@inproceedings{rei2023scaling,
    title = "Scaling up {C}omet{K}iwi: Unbabel-{IST} 2023 Submission for the Quality Estimation Shared Task",
    author = "Rei, Ricardo  and
      Guerreiro, Nuno M.  and
      Pombal, Jos{\~A}{\copyright}  and
      van Stigt, Daan  and
      Treviso, Marcos  and
      Coheur, Luisa  and
      C. de Souza, Jos{\'e} G.  and
      Martins, Andr{\'e}",
    editor = "Koehn, Philipp  and
      Haddow, Barry  and
      Kocmi, Tom  and
      Monz, Christof",
    booktitle = "Proceedings of the Eighth Conference on Machine Translation",
    month = dec,
    year = "2023",
    address = "Singapore",
    publisher = "Association for Computational Linguistics",
    url = "https://aclanthology.org/2023.wmt-1.73",
    doi = "10.18653/v1/2023.wmt-1.73",
    pages = "841--848",
}

@inproceedings{rei2024tower,
    title = "Tower v2: Unbabel-{IST} 2024 Submission for the General {MT} Shared Task",
    author = "Rei, Ricardo  and
      Pombal, Jose  and
      Guerreiro, Nuno M.  and
      Alves, Jo{\~a}o  and
      Martins, Pedro Henrique  and
      Fernandes, Patrick  and
      Wu, Helena  and
      Vaz, Tania  and
      Alves, Duarte  and
      Farajian, Amin  and
      Agrawal, Sweta  and
      Farinhas, Antonio  and
      C. De Souza, Jos{\'e} G.  and
      Martins, Andr{\'e}",
    editor = "Haddow, Barry  and
      Kocmi, Tom  and
      Koehn, Philipp  and
      Monz, Christof",
    booktitle = "Proceedings of the Ninth Conference on Machine Translation",
    month = nov,
    year = "2024",
    address = "Miami, Florida, USA",
    publisher = "Association for Computational Linguistics",
    url = "https://aclanthology.org/2024.wmt-1.12",
    doi = "10.18653/v1/2024.wmt-1.12",
    pages = "185--204",
}

@inproceedings{reimers2019sentence-bert,
    title = "Sentence-BERT: Sentence Embeddings using Siamese BERT-Networks",
    author = "Reimers, Nils and Gurevych, Iryna",
    booktitle = "Proceedings of the 2019 Conference on Empirical Methods in Natural Language Processing",
    month = "11",
    year = "2019",
    publisher = "Association for Computational Linguistics",
    url = "http://arxiv.org/abs/1908.10084",
}

@inproceedings{saichyshyna2023extension,
  title={Extension Multi30K: Multimodal dataset for integrated vision and language research in Ukrainian},
  author={Saichyshyna, Nataliia and Maksymenko, Daniil and Turuta, Oleksii and Yerokhin, Andriy and Babii, Andrii and Turuta, Olena},
  booktitle={Proceedings of the Second Ukrainian Natural Language Processing Workshop (UNLP)},
  pages={54--61},
  year={2023}
}

@misc{team2022NLLB,
author = {Team, NLLB and Costa-jussa, Marta and Cross, James and Çelebi, Onur and Elbayad, Maha and Heafield, Kenneth and Heffernan, Kevin and Kalbassi, Elahe and Lam, Janicec and Licht, Daniel and Maillard, Jean and Sun, Anna and Wang, Skyler and Wenzek, Guillaume and Youngblood, Al and Akula, Bapi and Barrault, Loïc and Gonzalez, Gabriel and Hansanti, Prangthip and Wang, Jeff},
year = {2022},
month = {07},
pages = {},
title = {No Language Left Behind: Scaling Human-Centered Machine Translation},
doi = {10.48550/arXiv.2207.04672}
}

@article{young2014image,
    title = "From image descriptions to visual denotations: New similarity metrics for semantic inference over event descriptions",
    author = "Young, Peter  and
      Lai, Alice  and
      Hodosh, Micah  and
      Hockenmaier, Julia",
    editor = "Lin, Dekang  and
      Collins, Michael  and
      Lee, Lillian",
    journal = "Transactions of the Association for Computational Linguistics",
    volume = "2",
    year = "2014",
    address = "Cambridge, MA",
    publisher = "MIT Press",
    url = "https://aclanthology.org/Q14-1006",
    doi = "10.1162/tacl_a_00166",
    pages = "67--78",
}

@inproceedings{zhu2023beyond,
    title = "Beyond Triplet: Leveraging the Most Data for Multimodal Machine Translation",
    author = "Zhu, Yaoming  and
      Sun, Zewei  and
      Cheng, Shanbo  and
      Huang, Luyang  and
      Wu, Liwei  and
      Wang, Mingxuan",
    editor = "Rogers, Anna  and
      Boyd-Graber, Jordan  and
      Okazaki, Naoaki",
    booktitle = "Findings of the Association for Computational Linguistics: ACL 2023",
    month = jul,
    year = "2023",
    address = "Toronto, Canada",
    publisher = "Association for Computational Linguistics",
    url = "https://aclanthology.org/2023.findings-acl.168",
    doi = "10.18653/v1/2023.findings-acl.168",
    pages = "2679--2697",
}

\end{document}